\documentclass[eat,twocolumn]{jmlr}

\usepackage{enumitem}
\usepackage{multirow}
\usepackage[export]{adjustbox}
\usepackage{float}
\usepackage[short]{optidef}
\usepackage{longtable}% for long tables
\usepackage{graphicx}
\usepackage{booktabs}
\usepackage[load-configurations=version-1]{siunitx} % newer version

 \usepackage{caption}
\theorembodyfont{\upshape}
\theoremheaderfont{\scshape}
\theorempostheader{:}
\theoremsep{\newline}

\DeclareMathOperator*{\argmin}{arg\,min}

\jmlrvolume{}
\firstpageno{1}
\jmlryear{2022}
\jmlrworkshop{Machine Learning for Health (ML4H) 2022}

\title[Checklists with continuous medical data]{Learning predictive checklists from continuous medical data}

\author{\Name{Yukti Makhija}
\Email{bb1190067@iitd.ac.in}\\
\addr Indian Institute of Technology, Delhi
\AND
\Name{Edward {de Brouwer}}
\Email{edward.debrouwer@gmail.com}\\
\addr ESAT-STADIUS, KU Leuven, Belgium
\AND
\Name{Rahul {G. Krishnan}}
\Email{rahulgk@cs.toronto.edu}\\
\addr University of Toronto, Vector Institute
}

\begin{document}

\maketitle

\begin{abstract}
%Successfull deployment of machine learning xmodel in clinical practice
%requires high standards of robustness and trustworthiness
%that often hinge on interpretability.
Checklists, while being only recently introduced in the medical domain,
have become highly popular in daily clinical practice due to their combined effectiveness
and great interpretability.
Checklists are usually designed by expert clinicians that manually collect and analyze available evidence.
However, the increasing quantity of available medical data is calling for a partially automated checklist design.
Recent works have taken a step in that direction by learning predictive checklists from categorical data.
In this work, we propose to extend this approach to accomodate learning checklists from continuous medical data using mixed-integer programming approach.
We show that this extension outperforms a range of explainable machine learning baselines on the prediction of sepsis from intensive care clinical trajectories.

\end{abstract}
\begin{keywords}
Checklists, Interpretability, Optimization, Concepts Learning.
\end{keywords}

\section{Introduction}
\label{sec:intro}

Recent years have seen a growing development of machine learning models
in the healthcare domain thanks to their impressive performance on a wide range of medical tasks \citep{davenport2019potential,esteva2019guide}.
However, despite the proliferation of architectures,
the adoption of machine learning models in clinical practice
remains a significant challenge~\citep{futoma2020myth,ahmad2018interpretable,ghassemi2020review,de2022predicting}. Indeed, ensuring the level robustness required for healthcare applications is difficult for complex models due to their inherent black box nature. Non-interpretable models make stress testing arduous and thus hamper the confidence required to deploy them in critical applications such as clinical practice.
Recent works have thus aimed at developing novel architectures that are both interpretable and retain most of the performance from their black box counterparts~\citep{ahmad2018interpretable}.

One such approach is learning medical checklists from available medical records.
Due to their simplicity and their ability to assist clinicians in complex situations, checklists have become increasingly popular in medical practice~\citep{haynes2009surgical}.
However, the design of such checklists is usually performed by expert clinicians, who manually collect evidence about the particular clinical problem of interest~\citep{hales2008development}.
As the number of available medical records grows, the manual collection of evidence becomes more tedious, bringing the need for partially automated design of medical checklists.
Recent works have taken a step in that direction by learning predictive checklists from boolean or categorical medical data~\citep{zhang2021learning}.
Nevertheless, many available clinical data, such as images or time series, are not categorical by nature, and therefore fall outside the limits of applicability of previous approaches.

In this work, we relax the previous categorical assumption and propose a novel approach to learning checklists from continuous-valued medical data. In particular, we propose a mixed-integer programming formulation
and show that our approach outperforms other interpretable baselines in predicting sepsis from critical care trajectories.

\iffalse
Contents:
\begin{itemize}
    \item The problem with interpretability in healthcare and the need to learn interpretable concepts for clinical discovery.
    \item Checklists. Why they are useful and widely adopted in clinical practice (maybe refer to the Checklist manifesto).
    \item Learning Checklists from data. Why it's desirable.
    \item Limitations of learning from boolean data
    \item Other data modalities (images, time series) have no clear way to be incorporated in checklists but are ubiquitous in medical datasets.
    \item Short description of our approach. Relaxes the boolean input assumption, mixed-integer and continuous program that allows to learn thresholds from data. (Paves the way towards differentiable approach).
\end{itemize}
\fi

\section{Problem setup}

Generally, we define a predictive checklist as a linear classifier that labels a patient as positive if $\mathbf{M}$ out of $\mathbf{N}$ rules are being satisfied, and negative otherwise.

Considering a medical dataset with $\mathbf{n}$ patients, $(\mathbf{X_i},\mathbf{y_i})$ for $\mathbf{i} \in [\mathbf{n}]$, we let $\mathbf{X_i}$ be the continuous or binary available medical variables and $\mathbf{y_i} \in \{0,1\}$ the label (\emph{e.g.} a positive or negative diagnosis). We further define  $\mathbf{I^+}$ and $\mathbf{I^-}$ as the list of indices of positive and negative patients respectively.

For a patient $i$, the checklist prediction is given by

\begin{align}
        \mathbf{\hat{y}_i = (w^T C(X_i) \geq M)}
\end{align},

where $C(X_i)$ is a vector of binary concepts or rules whose values are derived from the variables $\mathbf{X_i}$, $\mathbf{w} \in \{0,1\}^d $ are learnable binary weights and $M \in \mathbb{R}$ the threshold parameter. Letting $\mathbf{l^+}$ and $\mathbf{l^-}$ be the total number of misclassified positive and negative patients respectively, our goal is to learn the concepts, weights and threshold parameter such that the prediction minimize some distance $\mathcal{L}$ between the true labels and the predicted ones.

\begin{align*}
\tag{2}
    w^*,C^*,M^* =  \argmin_{w,C,M} \  \mathcal{L} (\mathbf{y},\hat{\mathbf{y}})
\end{align*}

\section{Methods}

Previous formulation of the learnable checklist  relying on binary input variables used $\mathbf{C(X_i)} = \mathbf{X_i}$~\citep{zhang2021learning}. However, this identity map cannot hold anymore if $\mathbf{X_i}$ is continuous. To address this limitation, we instead propose to define concepts based on a learnt threshold:

\begin{align}
\tag{3}
        \mathbf{C(X_i) = sign(X_i-t)}
\end{align}

The threshold for the $\mathbf{j^{th}}$ feature is denoted by $\mathbf{t_j \in \mathbb{R}}$. In contrast to previous approaches, our concepts are now \emph{learnable} functions of the input data. %We then propose the following mixed integer program (MIP) to compute the optimal checklist from continuous data.
The optimal checklist can then be found by solving the following mixed-integer program (MIP):

\begin{align}
\tag{4}
    \min_{\mathbf{w,z,M,t}} \mathbf{l^+ + \lambda l^- + \epsilon_N N + \epsilon_M M}
\end{align}

\textbf{s.t.}
\begin{equation}
    \tag{5a}
    \mathbf{A_jC_{i,j} > X_{ij} - t_j}{\ \ \ \ \ \ \ \mathbf{X_{i,j}>t_j}}
\end{equation}
\begin{equation}
    \tag{5b}
    \mathbf{A_jC_{i,j} < t_j - X_{i,j}}{\ \ \ \ \ \ \ \mathbf{X_{i,j} \leq t_j}}
\end{equation}
\begin{equation}
    \tag{6a}
    \mathbf{B_i z_i \geq M - w^T C_i} \ \ \ \ \ \ \ \ \mathbf{i \in I^+}
\end{equation}
\begin{equation}
    \tag{6b}
    \mathbf{B_i z_i \geq w^T C_i - M + 1}\ \ \ \mathbf{i \in I^-}
\end{equation}
\begin{equation}\notag
    % \tag{5a,b}
    \mathbf{l^+ = \sum_{\ i \in I^+}z_i} \ \ \ \ \ \ \ \ \mathbf{l^- = \sum_{\ i \in I^-}z_i}
\end{equation}
\begin{equation}\notag
    % \begin{lfalign}
        \mathbf{N = \sum_{j=1}^{d} w_j}
    % \end{lfalign}
    \ \ \ \ \ \ \ \mathbf{w_j \in \{0,1\}}
\end{equation}
\begin{equation}\notag
    \mathbf{C_i \in \{0,1\}^d} \ \ \ \ \ \ \ \ \ \ \ \ \mathbf{i \in [n]}
\end{equation}
\begin{equation}\notag
    \mathbf{M \leq N} \ \ \ \ \ \ \ \ \ \ \ \ \ \ \ \ \ \ \ \ \ \ \ \ \ \ \
\end{equation}

% \end{align}
Here, the objective minimizes the mistakes and based on the application, we can select an appropriate value of $\mathbf{\lambda}$ to balance between sensitivity and specificity. For instance, in multiple medical settings where a high specificity is required, a larger value of $\mathbf{\lambda}$ can be chosen to increase the cost of incorrect predictions for negative patients.

Big-M Constraints (5a) and (5b) are essential for learning appropriate thresholds and assign $\mathbf{C_{ij}}$ as 1 if $\mathbf{X_{ij}>t_j}$ and vice-versa. These constraints are dependent on a vector $\mathbf{A \geq |X_{max} - X_{min}|}$.

\iffalse
\section{Related works}

Several design strategies have been identified for creating effective medical checklists in the medical literature such as the creation of expert panels or the collection of pre-published guidelines \citep{hales2008development}.
Among these, our work aims at inferring checklists from data directly. In that regard, our approach extends the previous work of
\citet{zhang2021learning} who proposed an integer programming approach to learn optimal predictive checklists from boolean data. Our work also builds upon

\fi

\section{Experiments}

\subsection{Dataset}

% \begin{itemize}
    % \item \textbf{MNIST}

    % Since the proposed MIP method is generic, we first test it on a synthetic dataset generated using MNIST. A rule-set was defined based on which the instances were classified as 0/1, and the goal here is to recover this checklist using our program.
    % The image labels were randomly divided into groups of four and their concatenated one-hot encoding was generated. The $\mathbf{i^{th}}$ group forms the feature vector for that sample $\mathbf{X_i} \in\{0,1\}^{40}$. If at least three of the following conditions are met then the sample is given a positive label ($\mathbf{y_i = 1}$).
    % \newlist{todolist}{itemize}{2}
    % \setlist[todolist]{label=$\square$}
    % \begin{itemize}
    % \item \textbf{Rule-set ($\mathbf{M} = 3$, $\mathbf{N} = 17$)}

    % \begin{todolist}
    %     \item $\mathbf{Label\, 1} \in\{0,2,4,6,8\}$
    %     \item $\mathbf{Label\, 2} \in\{1,3,5,7,9\}$
    %     \item $\mathbf{Label\, 3} \in\{4,5,6\}$
    %     \item $\mathbf{Label\, 4} \in\{6,7,8,9\}$
    % \end{todolist}
    % \end{itemize}
    % The training data had 10500 samples total, out of which nearly 20\% were positive. Since our goal was to work with continuous data and learn real-valued thresholds, we replaced the 0s in the one-hot encoding with 0.2 and 1s with 0.8.

    % \item \textbf{PhysioNet}

    We used the PhysioNet 2019 Early Sepsis Prediction time series dataset~\citep{reyna2019early}, which was collected from the ICUs of three hospitals. It contains thirty-four non-static variables for which hourly data is available. These include both vital signs and laboratory values. Four static variables, including age, gender, duration, and starting point of the anomaly, are also present. We take the occurrence of sepsis in patients as the binary outcome variable.

    In the original training dataset, there were 32,268 patients, out of which only 8\% were positive. To reduce the imbalance, we created five subsets with 2200 patients in each fold, and about 37\% of patients had experienced sepsis.

    For each patient, we use mean, standard deviation and the last entry of each variable in the available clinical time series. We subsequently perform feature selection and only keep the top ten informative features based on a logistic regression.

    % We first trained a simple logistic regression using all the summary features and selected the top ten features based on the weights. This subset was then used for experimentation.

    %Interpretable summary functions defined in \cite{} were used to extract information from the non-static variables. We only use mean, standard deviation, and the last entry values for each feature.
% \end{itemize}

\begin{table*}[ht!]
\centering
\caption{Performance results of sepsis prediction on PhysioNet. We report accuracy, precision, recall as well as conciseness of the learnt checklist.}
\label{table:main_results}
\begin{adjustbox}{width=\textwidth}
    \begin{tabular}{p{0.3\linewidth}llllll}
    \toprule
    \textbf{Model} & \textbf{Accuracy} & \textbf{Precision} & \textbf{Recall} & \textbf{Specificity} & \textbf{N} & \textbf{M}\\ \midrule
    Dummy Classifier & 62.77 & 0 & 0 & - & - \\
    MLP (non-interpretable) & \textbf{64.96 ± 2.59} & 0.57 ± 0.05 & 0.48 ± 0.07 & 0.76 ± 0.06 &- & - \\
    Logistic Regression & 62.56 ± 1.65 & \textbf{0.62 ± 0.05} & 0.14 ± 0.04 & 0.94 ± 0.03 & - & - \\
    Unit Weighting & 58.28 ± 3.58 & 0.52 ± 0.09 & 0.44 ± 0.3 & 0.69 ± 0.25 & 9.6 ± 0.8 & 3.2 ± 1.16 \\
    SETS Checklist & 56.48 ± 7.88 & 0.52 ± 0.11 & \textbf{0.66 ± 0.30} & 0.49 ± 0.32 & 10 ± 0 & 6 ± 0.63 \\
    ILP mean thresholds & 62.99 ± 0.82 & 0.54 ± 0.087 & 0.12 ± 0.09 & 0.93 ± 0.32 & 4.4 ± 1.01 & 2.8 ± 0.75 \\
    \textbf{MIP (\textit{ours})} & 63.69 ± 2.44 & 0.56 ± 0.05 & \textbf{0.40 ± 0.08} & \textbf{0.79 ± 0.06} & 8 ± 1.09 & 3.6 ± 0.8 \\ \bottomrule
    \end{tabular}
    \end{adjustbox}
\end{table*}
\subsection{Models and baselines}

We compare our approach against several baselines such as a logistic regression and a classical multi-layer perceptron (MLP) as well as checklist-specific architectures like Unit weighting, SETS checklist and Integer Linear Program (ILP) with mean thresholds. Unit weighting is way to distill a pre-trained logistic regression into a checklist as also used in~\citet{zhang2021learning}. SETS checklists are
a novel baseline that we propose and consist of a modified logistic regression that incorporate a temperature parameter to induce binary weighting. ILP is the method proposed by \citet{zhang2021learning} to learn checklists from boolean data and is the most pertinent baseline to compare our approach with. To use this method, we binarized the data using mean thresholding. More details about the baselines are to be found in Appendix~\ref{app:baselines}.

% \begin{itemize}

    % \item \textbf{Unit weighting}

% \end{itemize}

\subsection{Predictive checklist performance}

In Table~\ref{table:main_results}, we report the performance of our approach and baselines on the PhysioNet dataset in terms of accuracy, precision, recall and conciseness of the checklist. We observe a significant improvement in recall when the thresholds are learnt, as opposed to the original ILP, where mean binarization is applied. This indicates that the minority class is being classified properly. Even though the best accuracy is achieved with an MLP, it comes at the cost of lower interpretability.

%All the major results from PhysioNet have been summarised in Table \ref{table:main_results}. %We first trained a simple logistic regression using all the summary features and selected the top ten features based on the weights. This subset was then used for experimentation.

%We saw a significant improvement in recall when the thresholds were learnt from our program, as opposed to the original ILP, where mean binarization is applied. This indicates that the minority class is being classified properly. Even though the best accuracy is achieved with an MLP, it comes at the cost of interpretability.

\subsection{Operating point comparison}

The balance between precision and recall in our checklist is tuned via the parameter $\mathbf{\lambda}$. Yet, each individual checklist leads to a single operating point. For finer assessment of our method, we compare its performance to a logistic regression evaluated at the precision and recall of our learnt checklist, as reported in Table~\ref{table:operating_points_results}.
%Then finding recall when precision for logistic regression is equal to the value obtained by solving our MIP. The reverse was done where the recall of a logistic regression was examined while keeping precision fixed.

% \begin{table}[h!]
% \begin{tabular}{|l|l|l|}
% \begin{tabularx}{0.5\textwidth}{@{}c L S[table-format=7.0]@{}}

\begin{table}[ht!]
\centering
\caption{Comparison of the Logistic Regression and MIP methods at different operating points, precision at recall = 0.403 (P@R=0.403) and recall at precision = 0.563 (R@P=0.563).}
\label{table:operating_points_results}
\begin{adjustbox}{width=\linewidth}
\begin{tabular}{lcc}
 & P@R=0.403 & R@P= 0.563 \\ \hline
\textbf{Logistic Regression} & 0.545 ± 0.052 & $\textbf{0.468 ± 0.23}$ \\ \hline
\textbf{MIP \textit{(ours)}} & $\textbf{0.563 ± 0.05}$ & 0.403 ± 0.08 \\ \hline
\end{tabular}
\end{adjustbox}
\end{table}

% \end{table}
The logistic regression achieves a lower precision value at the same recall level as our checklist. However, regarding the recall value at our checklist's precision score, the logistic regression is superior. We note that a checklist, due to its binary weights, has a strictly lower capacity\footnote{The capacity of the model should be understood here in terms of the VC-dimension.}, and is thus less expressive, than a logistic regression but results in a more practical and interpretable model.

%A plausible reason behind this is that in a checklist, all the items have the same weight and are considered equivalent. Logistic regression, on the other hand, can learn different real-valued weights for the features, thereby assigning them a different degree of importance. Compact binary rules and checklists are superior in terms of interpretability.

\subsection{Learning informative thresholds for clinical variables}

\begin{figure}[htb!]
\centering
\includegraphics[width=7cm, height=4cm]{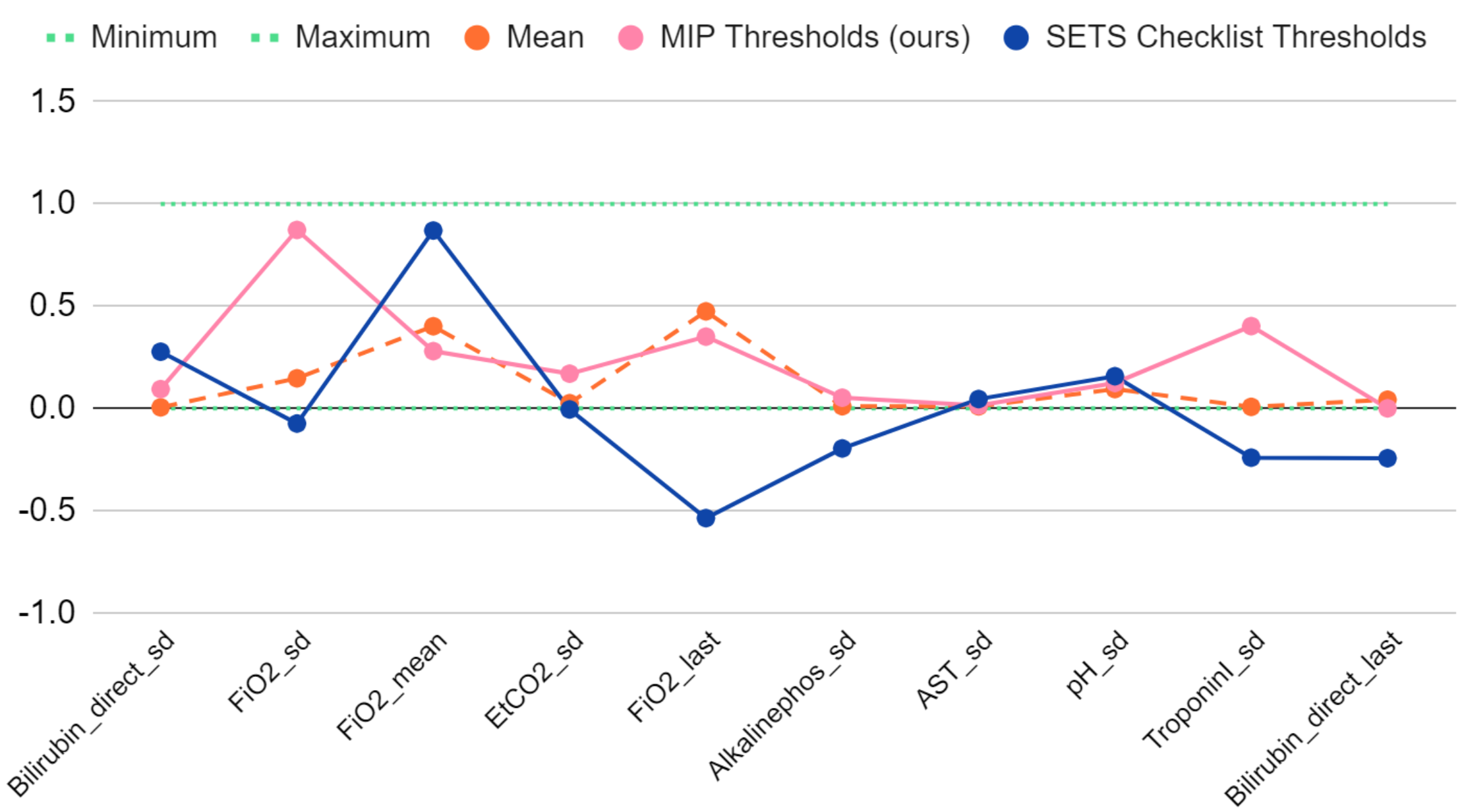}
\caption{Comparison of Thresholds Learnt by our MIP and SETS Checklists with the Mean}
\label{fig:my_label}
\end{figure}

To evaluate our MIP, we also compared the thresholds learnt with those obtained from the SETS checklist method and the mean values of the features. The results can be seen in Figure \ref{fig:my_label}, where we have normalized the features and the threshold values.

For the logistic regression, the thresholds for some features are falling outside their possible range. From the plot, we deduce that FiO2\_sd and Troponinl\_sd have considerably different thresholds from their mean values, while the remaining thresholds lie close to their means. Even though the model learns different thresholds only for a few features, this leads to a remarkable improvement in the recall, thereby reiterating the importance of our method and its ability to learn informative thresholds.

%\subsection{Results}
%For the synthetic dataset (created using MNIST), the MIP recovered the correct checklist, which led to a perfect performance on the training and testing datasets. Since the distribution of the digits was uniform across the four labels, our method learns 0.7999 as the threshold for all 40 features.

\iffalse
Compare Accuracy, Precision, Recall, N, M.
Add the results table.
\fi

\section{Conclusion and Future work}

From our experiments, it is clear that learning thresholds is beneficial, therefore demonstrating the usefulness of our approach in the continuous clinical data setting.
%Our method works went we receive raw data in tabulated format, and clinical parameters can be directly used as concepts.
Nevertheless, our approach still lacks the ability to process more complex data modalities, such as histopathological images, surgical videos, and multimodal datasets. To achieve this, we are planning to build a hybrid pipeline where we jointly extract relevant concepts from the data using deep learning models and embed our combinatorial solver as a layer in the overall architecture.

An emerging area of research in ML is the development of differentiation techniques for discrete optimizers, such as informative gradient approximation (\citet{berthet2020learning},\citet{paulus2021comboptnet},\citet{dalle2022learning}), introducing approximations to create continuous and soft versions of combinatorial algorithms(\citet{wang2019satnet}). Most of these have only been tested on synthetic datasets so far. We believe extending them to healthcare applications where a colossal amount of real-world data is available would yield fruitful results.

\section{Acknowledgments}

This research was supported by NSERC Discovery Award RGPIN-2022-04546 and a CIFAR AI Chair. Resources used in preparing this research were provided, in part, by the Province of Ontario, the Government of Canada through CIFAR, and companies sponsoring the Vector Institute. Edward De Brouwer was funded by a FWO-SB grant from the Flemish government. Yukti Makhija was funded by a research grant provided through the federal Pan-Canadian Artificial Intelligence Strategy.

\iffalse
\begin{itemize}
    \item Differentiability through the solver.
    \item MNIST with learnt concepts.
\end{itemize}
\fi

% \section{Conclusion}

% \begin{itemize}
%     \item Discuss current results (how learning the threshold is beneficial)
%     \item (Fairness criteria)
%     \item Future work: more complex concept learning
%     \item Difficulty of the trade-off between performance and interpretability.
% \end{itemize}

%\acks{Acknowledgements go here.}
\newpage

\bibliography{pmlr-sample}

\appendix

\section{Baselines details}\label{app:baselines}

\textbf{Unit weighting}

    Following~\citet{zhang2021learning}, we use unit weighting to obtain a valid checklist from a logistic regression models trained on binarized data. The dataset is binarized using the mean of the features as the threshold. Features with positive weights and complements of those with negative weights are added to the checklist. A hyperparameter $\mathbf{\beta}$ is used to filter out the features with values between $\mathbf{[-\beta, \beta]}$. Then the optimal value of $\mathbf{M}$ is picked by training logistic regression models on the binarized dataset and setting different values of $\mathbf{M \in [N]}$.\\

    % \item \textbf{SETS checklist}
\textbf{SETS checklist}

    This approach learns a threshold vector, $\mathbf{\phi}$ for each feature as a parameter while training the logistic regression.
    \begin{equation}
    \tag{7}
        \mathbf{\sigma(\frac{(X_i - \phi)}{\tau})}
    \end{equation}
    $\mathbf{\tau}$ is a temperature parameter for the sigmoid ($\mathbf{\sigma}$) function. We use these thresholds to binarize the data and perform unit weighting to generate the checklists.\\
    % \item \textbf{Integer Linear Program (ILP) with mean thresholds}

\textbf{Integer Linear Program (ILP) with mean thresholds}

    We re-implemented the ILP from \cite{zhang2021learning} and solved it using Gurobi. The continuous dataset was binarized through mean thresholding. This is the most pertinent baseline to compare our approach against.

\end{document}